\documentclass[journal]{IEEEtran}
\pdfoutput=1
\ifCLASSINFOpdf
\else
\fi

\usepackage{float}
\usepackage{mathtools}
\usepackage{amssymb}

\usepackage{hyperref} 
\hypersetup{	breaklinks,
	colorlinks=true,       
    linkcolor=blue,          
    citecolor=green,        
    filecolor=magenta,      
    urlcolor=black,           
  pdfauthor={Guillermo Gallego, Christian Forster, Elias Mueggler, Davide Scaramuzza},%
  pdftitle={Event-based Camera Pose Tracking using a Generative Event Model},%
  pdfsubject={Computer Vision, Robotics, Image processing, Neuromorphic sensors},%
  pdfkeywords={Event-based, Dynamic Vision Sensor, generative model, spiking model, robot localization, pose tracking, Kalman filter, light intensity, optic flow}%
}

\usepackage{soul} 

\usepackage{balance} 

\providecommand{\tabularnewline}{\\}
\floatstyle{ruled}
\newfloat{algorithm}{tbp}{loa}
\providecommand{\algorithmname}{Algorithm}
\floatname{algorithm}{\protect\algorithmname}

\global\long\def\R{\mathbb{R}}
\global\long\def\X{\mathbf{X}}
\global\long\def\x{\mathbf{x}}

\global\long\def\f{\mathbf{f}}
\global\long\def\h{\mathbf{h}}
\global\long\def\g{\mathbf{g}}
\global\long\def\ba{\mathbf{a}}
\global\long\def\bb{\mathbf{b}}

\global\long\def\bu{\mathbf{u}}
\global\long\def\bv{\mathbf{v}}
\global\long\def\bw{\mathbf{w}}
\global\long\def\br{\mathbf{r}}
\global\long\def\q{\mathbf{q}}
\global\long\def\z{\mathbf{z}}

\global\long\def\Rot{\mathtt{R}}
\global\long\def\bt{\mathbf{t}}

\global\long\def\linvel{\bv}
\global\long\def\bmu{\boldsymbol{\mu}}
\global\long\def\bnu{\boldsymbol{\nu}}
\global\long\def\bxi{\boldsymbol{\xi}}
\global\long\def\eeta{\boldsymbol{\eta}}
\global\long\def\bomega{\boldsymbol{\omega}}
\global\long\def\angvel{\boldsymbol{\omega}}

\global\long\def\linvelNoise{\mathbf{V}}
\global\long\def\angvelNoise{\boldsymbol{\Omega}}
\global\long\def\Cov{Q}
\global\long\def\prtl#1#2{\frac{\partial#1}{\partial#2}}

\global\long\def\inner#1#2{\left<#1,#2\right>}

\global\long\def\cM{\mathcal{M}}
\global\long\def\cN{\mathcal{N}}
\global\long\def\bp{\mathbf{p}}

\global\long\def\Int{I}
\global\long\def\IntS{I^{S}}
\global\long\def\diff#1#2{\frac{d#1}{d#2}}
\global\long\def\mB{\mathtt{B}}
\global\long\def\bel{\textit{bel}}
\global\long\def\sgn{\text{sgn}}

\begin{document}

%
\title{Event-based Camera Pose Tracking using a Generative Event Model}


\author{Guillermo~Gallego, Christian~Forster, Elias~Mueggler, Davide~Scaramuzza
\thanks{The authors are with the Robotics and Perception Group, Department of Informatics, University of Zurich, Zurich 8050, Switzerland. \href{http://rpg.ifi.uzh.ch}{http://rpg.ifi.uzh.ch} 
e-mail: guillermo.gallego@ifi.uzh.ch.}
\thanks{This research was supported by the National Centre of Competence in Research (NCCR) Robotics, the ERC-SNSF Starting Grant, and Google Faculty Research Award.}
}

\maketitle

\begin{abstract}
Event-based vision sensors mimic the operation of biological retina 
and they represent a major paradigm shift from traditional cameras.
Instead of providing frames of intensity measurements synchronously, at artificially chosen rates, 
event-based cameras provide information on brightness changes asynchronously, when they occur.
Such non-redundant pieces of information are called ``events''. 
These sensors overcome some of the limitations of traditional cameras (response time, bandwidth and dynamic range) but require new methods to deal with the data they output.
We tackle the problem of event-based camera localization in a known environment, without additional sensing, using a probabilistic generative event model in a Bayesian filtering framework.
Our main contribution is the design of the likelihood function used in the filter to process the observed events.
Based on the physical characteristics of the sensor and on empirical evidence of the Gaussian-like distribution of spiked events with respect to the brightness change, we propose to use the contrast residual as a measure of how well the estimated pose of the event-based camera and the environment explain the observed events.
The filter allows for localization in the general case of six degrees-of-freedom motions. 
\end{abstract}



\section{Introduction}
\label{sec:intro}
Recently, event-based cameras such as the Dynamic Vision Sensor (DVS)~\cite{Lichtsteiner08ssc} have attracted a lot of attention from both the robotics and vision communities~\cite{Weikersdorfer12robio,Weikersdorfer13icvs,Benosman14nn, Censi14icra,Mueggler14iros,Cook11IJCNN,Kim14bmvc,Weikersdorfer2014icra,Orchard2015}.
These bio-inspired sensors overcome some of the limitations of traditional image sensors: they respond very quickly (within microseconds) to brightness changes, have very high dynamic range (120 dB compared to 60 dB of standard cameras), and require low bandwidth~\cite{Lichtsteiner08ssc}.
Hence, they are very promising sensors for high-speed visual applications
in challenging scenes with large brightness contrast.
However, the output of these cameras (a stream of events) is fundamentally different from that of traditional ones, and so a paradigm shift is required to design algorithms that exploit the potential of these vision sensors.
Examples of such emerging event-based algorithms are: event-based optical flow~\cite{Benosman14nn}, 
visual odometry~\cite{Censi14icra},
localization~\cite{Weikersdorfer12robio,Mueggler14iros}, 
Simultaneous Localization and Mapping (SLAM)~\cite{Weikersdorfer13icvs,Weikersdorfer2014icra},
mosaicing~\cite{Cook11IJCNN,Kim14bmvc}, 
object recognition~\cite{Orchard2015}, etc.

We address the localization problem of a moving event-based camera in a known environment.
One of the first works in this respect is~\cite{Weikersdorfer12robio}, where a particle-filter system that is limited to planar motions and 2-D maps was introduced. 
In the experiments, they used an upward-looking DVS mounted on a ground robot moving at low speed. 
The provided map used for navigation consisted of line segments on the ceiling.
In~\cite{Censi14icra}, a probabilistic filtering approach was designed to localize a DVS moving on a plane with respect to the temporally closest pair of frames provided by an additional RGB-D camera attached to the DVS. 
An algorithm to track the 6-DOF pose of the DVS with no additional sensing
during high-speed maneuvers was given in~\cite{Mueggler14iros}. 
They used a map consisting of the edges of a black square of known size and minimized the event-to-line reprojection distance to estimate the DVS pose.

We propose an \emph{implicit} Extended Kalman Filter (EKF) approach~\cite{Thrun05book} to localize the DVS with respect to a given dense map of the 3-D scene (consisting of geometric and photometric information) without additional sensing (as in~\cite{Weikersdorfer12robio,Mueggler14iros,Kim14bmvc}), just using the information contained in the event stream.
The map is not constrained to consist only of lines, thus it is more general than those in~\cite{Weikersdorfer12robio,Mueggler14iros}, and it is also richer in brightness changes than the barcoded scenes in~\cite{Censi14icra}. 
We allow for localization in the general case of \mbox{6-DOF} motion of the DVS and design the filter accordingly. 
Our main contribution pertains to the design of the likelihood function used in the correction step of the EKF to process the observed events (Section~\ref{sub:Details-MeasEq}), by measuring how well the system state (DVS pose and velocity) and the map explain an event from the DVS using a contrast residual.
To do so, we first derive a simple yet compelling model for event generation (Section~\ref{sec:SceneModeling}). 
The technique is demonstrated on synthetic and real data in Section~\ref{sec:experiments}.

\section{Dynamic Vision Sensor (DVS): \\generative event model}
\label{sec:dvsDescription}

\begin{figure*}
\begin{tabular}{ccc}
\includegraphics[width=0.27\linewidth]{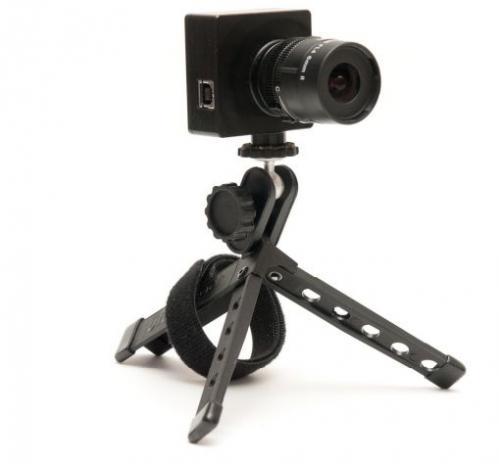} &
\includegraphics[width=0.33\linewidth]{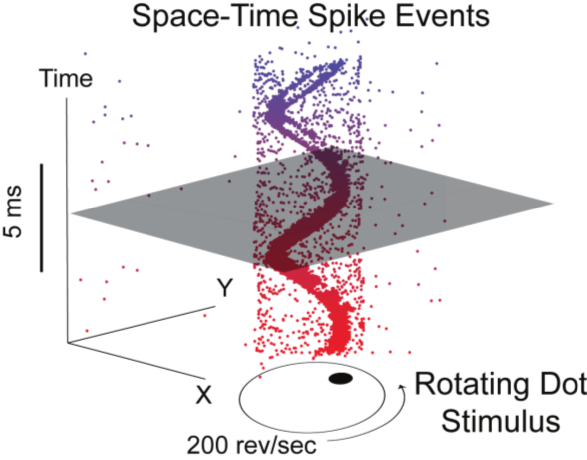} &
\includegraphics[width=0.33\linewidth]{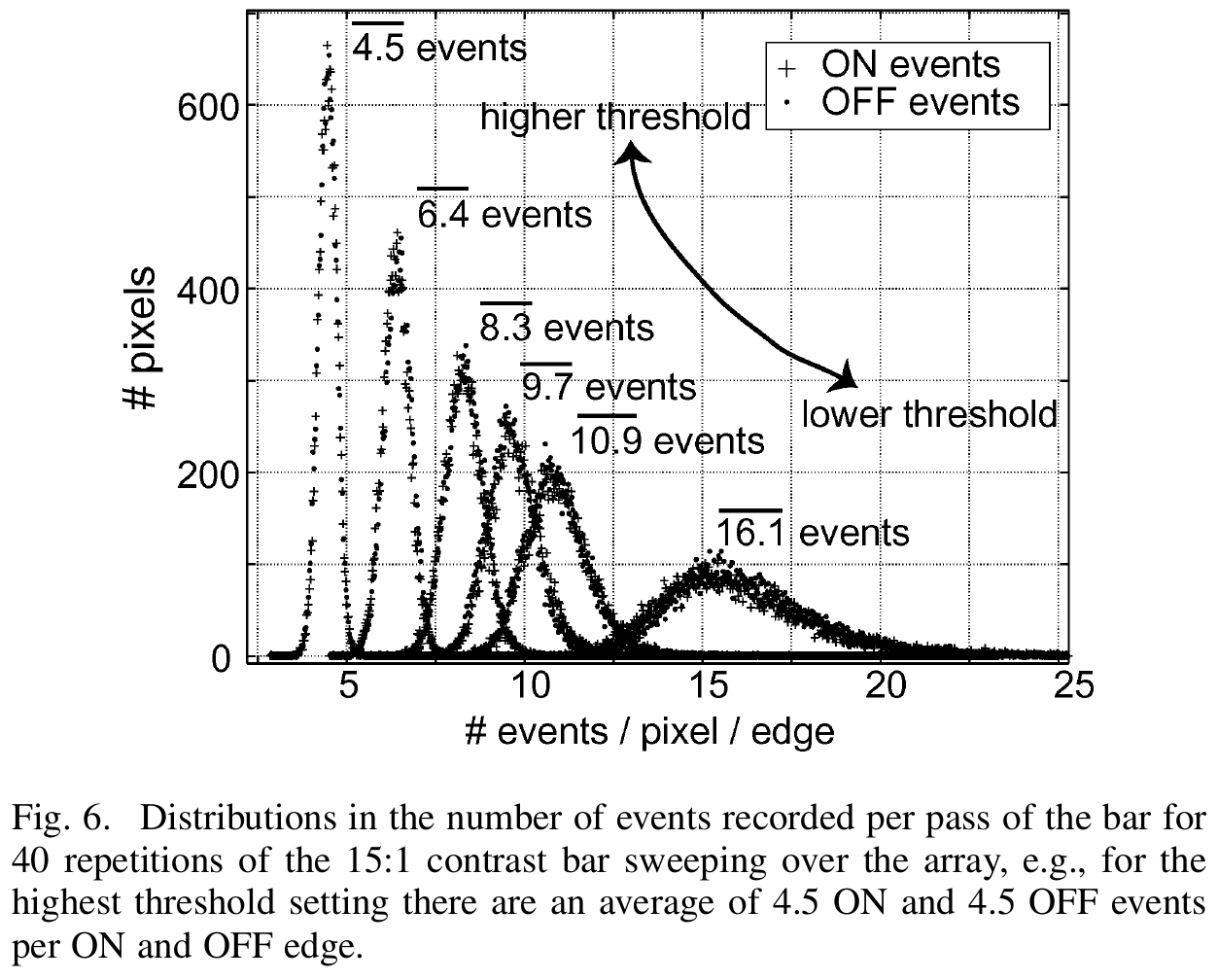}\\
(a)&(b)&(c)
\end{tabular}
\caption{(a) The Dynamic Vision Sensor (DVS). 
(b) Space-time visualization of the output of a DVS viewing a rotating dot. 
Colored dots mark individual events. Event polarity is not displayed. Noise is visible by isolated points that are not part of the spiral.
(c) The contrast of the DVS events empirically follows a unimodal distribution (e.g. Gaussian-like) centered at a selected threshold $C = |\Delta \log(I)|$ (six threshold settings are shown). 
Images (b) and (c) are courtesy of~\cite{Lichtsteiner08ssc}.
}
\label{fig:DVSContrast2008Composite}
\end{figure*}

In contrast to standard cameras, which acquire full frames at fixed rates, event-based vision sensors such as the DVS (Fig.~\ref{fig:DVSContrast2008Composite}a) have independent pixels that spike events at local relative brightness changes in continuous time.
A visualization of the output of the DVS is shown in Fig.~\ref{fig:DVSContrast2008Composite}b. 
Events are time-stamped with microsecond resolution and transmitted asynchronously at the time they occur. 
Each event is a tuple $e_{k}=\langle x_{k},y_{k},t_{k},p_{k}\rangle$, where $x_{k},y_{k}$ are pixel coordinates of the event, $t_{k}$ is its time-stamp, and $p_{k}=\pm1$ is its polarity (sign of the brightness change). 
The sensor's spatial resolution is limited\footnote{A new generation of event-based sensors with VGA resolution ($640 \times 480$) is being developed by the group~\cite{Lichtsteiner08ssc}.} ($128 \times 128$ pixels), but its 120~dB dynamic range notably exceeds the 60 dB of high-quality traditional image sensors.

\label{sub:Generative-model}
Next, we provide a generative event model for the DVS using a principled derivation of the equations that characterize an \emph{ideal} sensor.
The event model combines several hypothesis (constant brightness, temporal persistence, etc.) with particular characteristics of the DVS. 
The model is at the heart of data assimilation in our filtering approach for DVS localization.

\subsection{Scene modeling}\label{sec:SceneModeling}

Assume that objects in the 3-D world are represented by a surface $S$ with geometric and radiometric properties. 
Typically, objects are described by a mesh or depth map and a corresponding intensity (i.e., ``texture'') function (in a Lambertian context). 

The DVS has the same optics as traditional perspective cameras, therefore, standard models (e.g., pinhole) apply. 
In camera coordinates, the projection operation is described by $\bu=\pi(\X)$, mapping a 3-D point $\X=(X,Y,Z)^{\top}$ into the image point $\bu=(u,v)^{\top}$.

Assume a simplified radiance model where each point on the surface $S$ has an intensity, $\IntS:S\to\R$, and this is the value observed by the DVS to trigger events, that is, the intensity at the image plane corresponds to the intensity defined on the surface: $\Int(\bu)\doteq\IntS(\X)$ for 3-D points $\X$ visible from the DVS. 
Hence, the image plane parametrizes both the image $I$ and the surface $S$ (geometric and photometric properties). 

\subsection{3-D motion and apparent (2-D) motion}\label{sec:ApparentMotion}

The motion of a moving camera is described by a smooth trajectory in the space of Euclidean transformations,  $SE(3)$. 
Let the relative motion between the viewing camera and the scene be described, in the camera coordinate frame, by
\begin{equation}
\diff{\X}t\equiv\dot{\X}(t)=-\widehat{\bomega}(t)\X(t)-\bv(t)
\label{eq:MotionThreeDim}
\end{equation}
where $\bomega$ and $\bv$ are body angular and linear velocities, respectively, and $\widehat{\bomega}$ is the cross-product matrix: $\widehat{\ba}\bb=\ba\times\bb\;\forall\ba,\bb$. 
The corresponding apparent motion of the 3-D point $\X$ is described by the velocity of the image point $\bu$, which comprises the \emph{image motion field}. 
Specifically, the equation that relates surface velocity (in the camera frame) to feature velocity (in normalized coordinates) is (see, e.g.~\cite{Chaumette08handbook}, \cite[Eq. 5.87]{Ma04book}), dropping the $t$ notation:
\begin{equation}
\diff{\bu}t\equiv\dot{\bu}=\mB\,\bxi,
\label{eq:ImageMotionFieldPlanar}
\end{equation}
where twist coordinates $\bxi=(\bv^{\top},\bomega^{\top})^{\top}$ encode the relative motion
and
\begin{equation}
\mB=\left(\begin{array}{cccccc}
\!\!\!-Z^{-1} & \!\!\!0 & \!\!uZ^{-1} & \!\!uv & \!\!-(1+u^{2}) & \!\!v\\
\!\!\!0 & \!\!\!-Z^{-1} & \!\!vZ^{-1} & \!\!1+v^{2} & \!\!-uv & \!\!-u
\end{array}\right)
\end{equation}
is called \emph{interaction matrix}, \emph{image Jacobian} matrix for a point feature, or \emph{feature sensitivity matrix}~\cite{Chaumette08handbook}, \cite[p. 460-462]{corke2011robotics}.
Typically, the surface is assumed to admit a depth map representation with respect to the camera, and so the depth of the 3-D point is parametrized in the image plane, $Z\equiv Z(u,v)$. 
Consequently, $\mB(u,v)$ is just a function of the surface and the image point. 
The motion field has two separate components for translation and rotation.

\subsection{Deterministic generative event model}

The standard hypothesis in measuring image motion is that the intensity structure of local time-varying image regions are approximately constant under motion for at least a short duration (temporal persistence).
Formally, if $\tilde{I}(\bu,t)$ is the space-time image intensity function measured by the DVS, the total derivative $d\tilde{I}/dt$ vanishes for those trajectories $\bu(t)$ of constant intensity values, $\tilde{I}(\bu(t),t)=\text{const}$, that is,
\begin{equation}
\diff{\tilde{I}}t=\inner{\nabla_{\bu}\tilde{I}}{\dot{\bu}}+\prtl{\tilde{I}}t=0,\label{eq:ConstantBrightnessTotalDeriv}
\end{equation}
where $\inner{\cdot}{\cdot}$ is the dot product, $\nabla_{\bu}\tilde{I}=\bigl(\prtl{\tilde{I}}u,\prtl{\tilde{I}}v\bigr)^{\top}$
are the first partial derivatives with respect to spatial coordinates and $\dot{\bu}=\left(\dot{u},\dot{v}\right)^{\top}$ is the motion field. 

The DVS senses brightness logarithmically\footnote{Using the chain rule it is easy to verify that, if $I\neq0$, both conditions $dI/dt=0$ and $d\tilde{I}/dt=0$ are equivalent.}: $\tilde{I}=\log(I)$, 
and it generates an event at a location $\bu$ if the amount of intensity (grey level) change $\Delta\log(I)$ during an interval $\Delta t$ (the time since the previous event at the same location), i.e., the \emph{contrast} 
\begin{equation}
\Delta\log(I)\approx\prtl{\log(\Int)}t\Delta t\stackrel{\eqref{eq:ConstantBrightnessTotalDeriv}}{=}-\inner{\nabla_{\bu}\log(\Int)}{\dot{\bu}\Delta t}\label{eq:ContrastDef}
\end{equation}
is larger than a threshold $C$~\cite{Lichtsteiner08ssc,Censi14icra} (typically 10-15\% relative brightness change):
\begin{equation}
\left|\Delta\log(I)\right|\approx\left|\inner{\nabla_{\bu}\log(\Int)}{\dot{\bu}\Delta t}\right|\geq C.\label{eq:IncrementLogI_PlanarPerspProj}
\end{equation}
Incorporating polarity, if the contrast $\Delta\log(I)\geq C$, a positive event ($p_{k}=+1$) is generated; if $\Delta\log(I)\leq-C$, a negative event ($p_{k}=-1$) is generated; otherwise, no event is fired. 

\subsection{Probabilistic generative event model}

Equation~\eqref{eq:IncrementLogI_PlanarPerspProj} is a hard decision model for the generation of events. 
A more realistic one takes into account sensor noise and manufacturing mismatches, yielding a soft decision represented by a smooth probability function. 
A characterization of the corresponding probability density averaged over all DVS pixels is shown in Fig. 6 of~\cite{Lichtsteiner08ssc} (see Fig.~\ref{fig:DVSContrast2008Composite}c), suggesting a unimodal Gaussian-like distribution, for which they measure its standard deviation as a function of the threshold $C$. 
This probabilistic generative event model can be included in a Bayesian filtering approach to process the events, 
as shown in the next section, where we adopt the simple yet powerful filter given by the Extended Kalman Filter (EKF), which assumes Gaussian probability distributions to keep a compact and manageable representation of the posterior probability of the DVS pose and velocity.

\section{Bayesian filtering approach}
\label{sec:StateSpace}

\subsection{State-space design}

In the popular Bayesian inference framework given by the EKF~\cite{Thrun05book} we can formulate the DVS localization problem with respect to a map $\cM$ as that of estimating the state of a system defined by its state-space representation (state and measurement equations). 

\label{sec:detailsStateEq}

The state equation is a non-linear function $\f$ of the state and the process noise
\begin{equation}
\x_{n}=\f(\x_{n-1},\bw_{n}),\quad\text{with}\quad
\x = (\bt^\top,\br^\top,\linvel^\top,\angvel^\top)^{\top}.
\label{eq:StateEq}
\end{equation}
As usual, subscripts $\{n-1,n\}$ denote temporal references.
The process noise $\bw_{n}$ is not additive and it is assumed to be zero-mean multivariate Gaussian distributed with covariance $Q_{n}^{\bw}$. 
The state vector describes the DVS pose (position and orientation) and its velocity: 
$\bt$ is the position of the optical center of the DVS, in world coordinates; 
$\br$ is the rotation vector parametrizing the orientation of the DVS by means of the exponential coordinates (as in the filter proposed by~\cite{Chiuso2002}) of the rotation matrix from the world to the camera frame, $\Rot=\exp(\widehat{\br})$;
and the linear and angular velocities~\eqref{eq:MotionThreeDim} ($\linvel$, $\angvel$) are given in world and camera (body) coordinates, respectively.

We chose the motion model $\f$ given by the constant velocity model, which is typical of SLAM approaches~\cite{DavisonICCV03}.
This accounts for general smooth motions of the DVS. 
By integration of the continuous motion over a time interval\footnote{Here $\Delta t$ is the time between prediction steps in the EKF, which may or may not coincide with the time between events at the same location in~\eqref{eq:ContrastDef} depending on whether events are processed in packets or individually.} $\Delta t=t_{n}-t_{n-1}$ 
and discretization, \eqref{eq:StateEq} becomes
\begin{equation}
\begin{cases}
\bt_{n} & =\bt_{n-1}+(\linvel_{n-1}+\linvelNoise)\Delta t,\\
\br_{n} & =\left(\log\left(\exp\left((\angvel_{n-1}+\angvelNoise)^{\wedge}\Delta t\right)\exp\left(\br_{n-1}\right)\right)\right)^{\vee},\\
\linvel_{n} & =\linvel_{n-1}+\linvelNoise,\\
\angvel_{n} & =\angvel_{n-1}+\angvelNoise,
\end{cases}\label{eq:StateEqPartitioned}
\end{equation}
where the noise is $\bw_{n}=(\linvelNoise^\top, \angvelNoise^\top)^\top$. 
The $\log$ and $\exp$ operators refer to the rotation group, $SO(3)$.
$\Delta \Rot=\exp\left((\angvel_{n-1}+\angvelNoise)^{\wedge}\Delta t\right)$ is the incremental rotation of angle $\theta=\|\angvel_{n-1}+\angvelNoise\|\Delta t$ around the axis defined by vector $(\angvel_{n-1}+\angvelNoise)$,
$\bu^{\wedge}$ is the cross-product matrix associated to a 3-vector $\bu$, and $S^{\vee}$ is the 3-vector associated to a $3\times3$ skew-symmetric matrix~$S$.

\subsection{Implicit measurement equation}\label{sub:Details-MeasEq}

In the standard EKF, the likelihood is specified by an equation $\z_{n}=\h(\x_{n})+\eeta_{n},$
where observations $\z_{n}$ are explicitly written in terms of the state and the observation noise~$\eeta_{n}$. 
This is the formulation used in classical visual localization and SLAM, where $\z_{n}$ consists of the image coordinates of sensed map landmarks, and $\h$ predicts the observations by using the camera model to project the landmarks.
This design choice implies Gaussian image coordinate noise, and it may also be applied to DVS localization~\cite{Weikersdorfer12robio}.
However, it does not take into account the generative event model (such as~\eqref{eq:IncrementLogI_PlanarPerspProj}).
In a different (non-localization) context, an alternative approach is given in~\cite{Kim14bmvc} to estimate the intensity gradient at each pixel: $\z_{n}$ consists of event rates and a generative model is used to write such explicit dependency. 
This design choice implies that the temporal (event-rate) noise is Gaussian, which is an arbitrary choice.

We depart from the previous explicit models (spatial or temporal measurements) and propose an implicit measurement equation 
\begin{equation}
\q(\z_{n},\x_{n})=\mathbf{0}\label{eq:MeasurementEqImplicit}
\end{equation}
to quantify how well the event generation model~\eqref{eq:IncrementLogI_PlanarPerspProj} is satisfied.
This leads to an \emph{implicit} EKF~\cite{Zhang92IEKF,Soatto93IEKF}.
Our design choice assumes that the deviations of the contrast from the nominal one that fires events is Gaussian, which Fig.~\ref{fig:DVSContrast2008Composite}c suggests to be. 
A similar unimodal density function is given in~\cite{Kim14bmvc} only for the correction step of rotation tracking. 

\begin{figure*}
\begin{center}
\begin{tabular}{cccc}
\includegraphics[height=35mm]{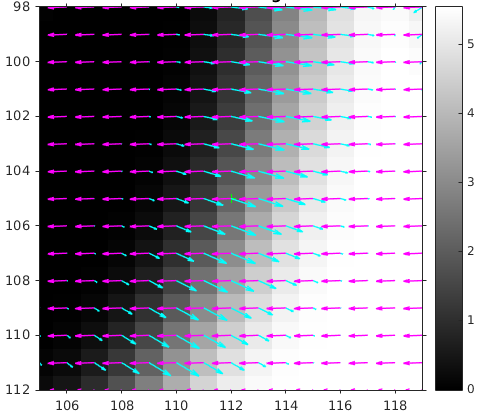}&
\includegraphics[height=35mm]{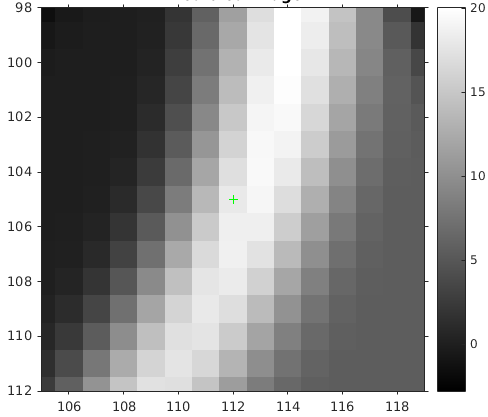}&
\includegraphics[height=35mm]{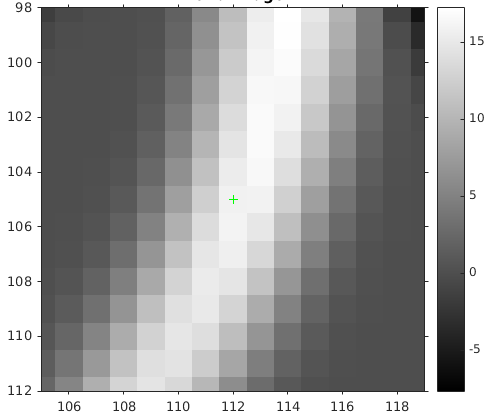}&
\includegraphics[height=35mm]{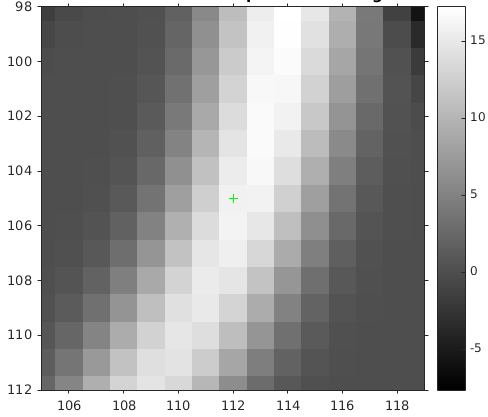}\\[3mm]
\includegraphics[height=35mm]{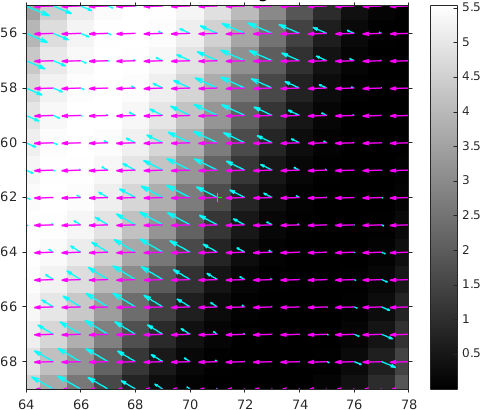}&
\includegraphics[height=35mm]{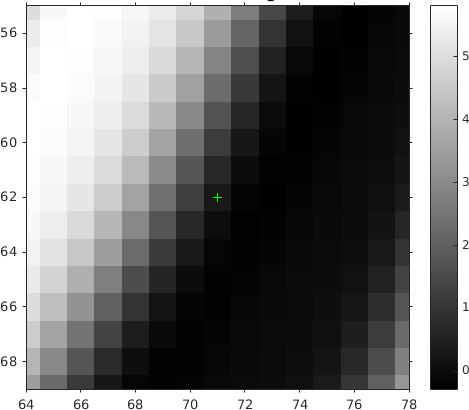}&
\includegraphics[height=35mm]{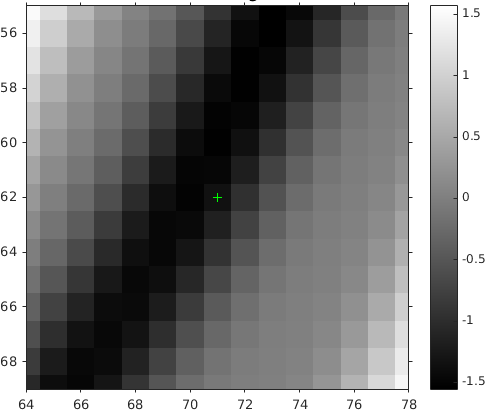}&
\includegraphics[height=35mm]{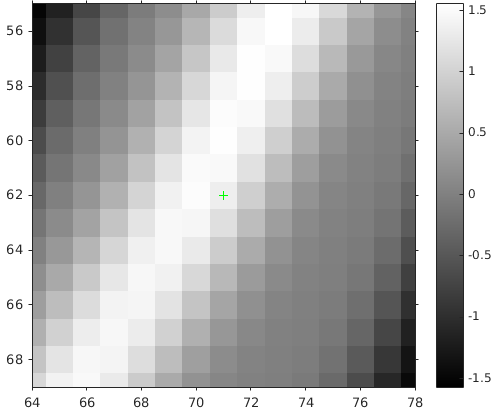}\\
(a)&(b)&(c)&(d)
\end{tabular}
\end{center}
\caption{Neighborhood of an event $e_n=\langle x_n,y_n,p_n,t_n \rangle$ triggered by a moving edge. 
The DVS is moving horizontally to the right (positive $X$ direction).
Top row: positive event (dark-to-bright transition).
Bottom row: negative event (bright-to-dark transition).
(a) Rendering of the map on the DVS image plane, $\tilde{I}(t_n-\Delta t_n)$; the event $\bp_n = (x_n,y_n)^\top$ is at the center of the patch. 
The motion field $\dot{\bu}$ (magenta vectors) points toward the negative $X$ direction.
The image gradient $\g=\nabla_\bu \tilde{I}(t_n-\Delta t_n)$ (perpendicular to the edge) is displayed with cyan vectors.
(b) Predicted neighborhood $\tilde{I}(t_n)\approx \tilde{I}(t_n-\Delta t_n) + \Delta \tilde{I}$.
(c) Constrast $\Delta \tilde{I} \approx -\inner{\g}{\dot{\bu}}\Delta t_n$.
(d) The implicit measurement function $\q$ in~\eqref{eq:ImplicitMeasDef} has the same shape as the absolute contrast, $|\Delta \tilde{I}| \approx -p_n \inner{\g}{\dot{\bu}}\Delta t_n$, 
which defines the likelihood that the event was triggered.}
\label{fig:DVSContrastAtPoint}
\end{figure*}
\begin{figure*}
\begin{center}
\begin{tabular}{cccc}
\includegraphics[height=35mm]{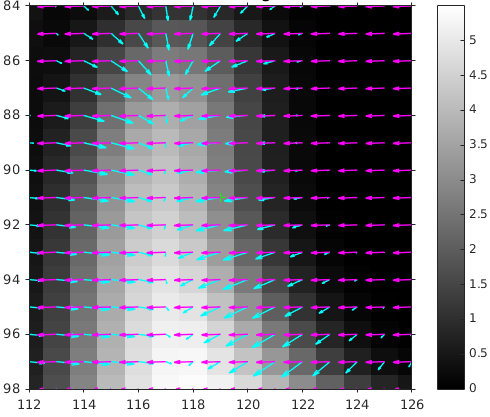}&
\includegraphics[height=35mm]{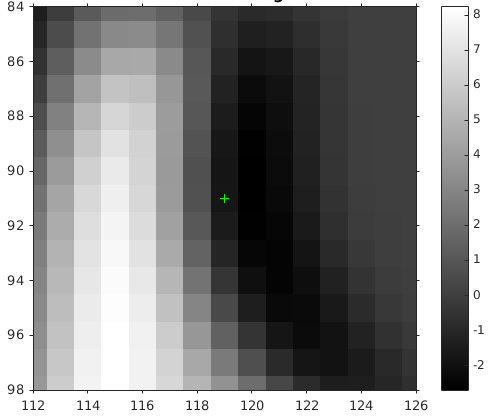}&
\includegraphics[height=35mm]{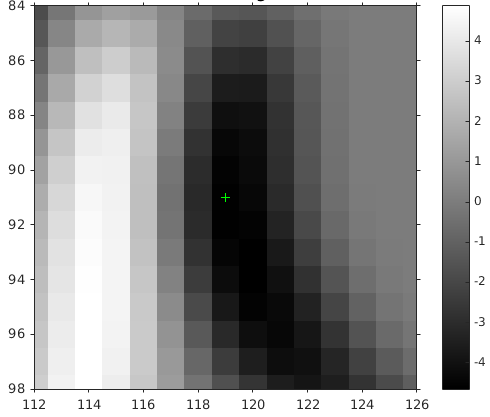}&
\includegraphics[height=35mm]{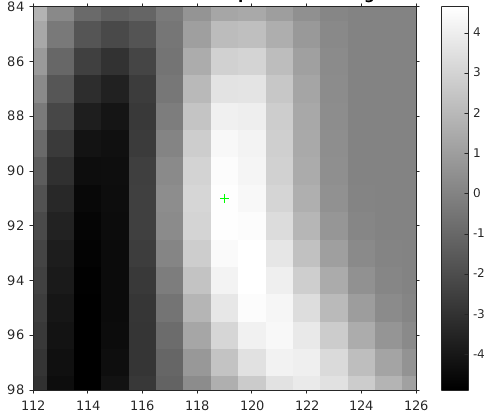}\\[3mm]
\includegraphics[height=35mm]{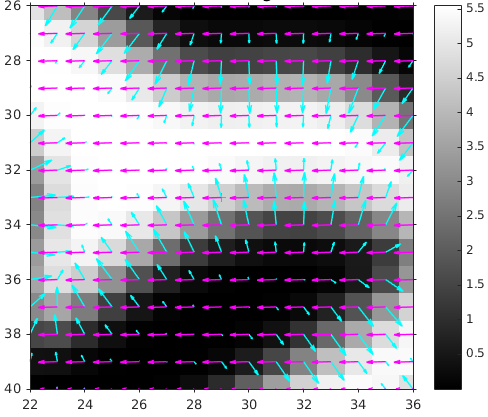}&
\includegraphics[height=35mm]{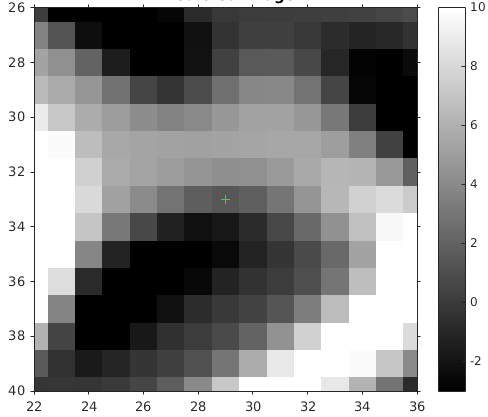}&
\includegraphics[height=35mm]{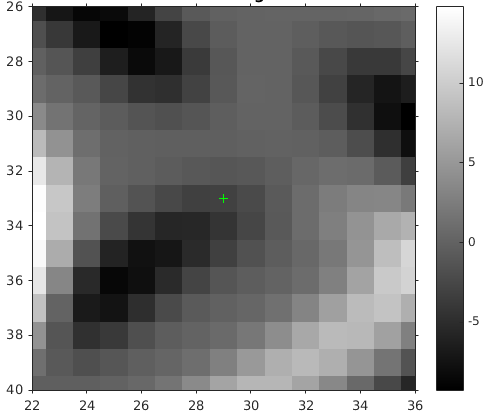}&
\includegraphics[height=35mm]{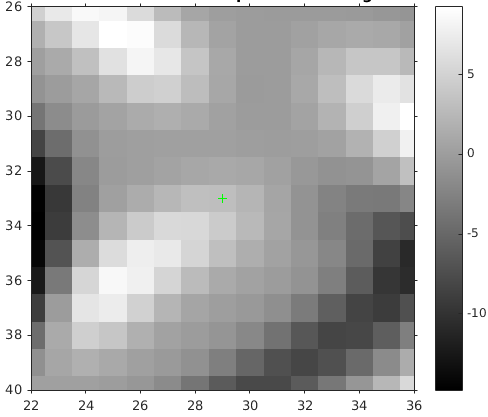}\\
(a)&(b)&(c)&(d)
\end{tabular}
\end{center}
\caption{Neighborhood of an event triggered by a moving edge.
Same notation as in Fig.~\ref{fig:DVSContrastAtPoint}.
Top row: at the event location, the image gradient $\g$ is parallel to the motion field $\dot{\bu}$.
Bottom~row: $\g$ almost perpendicular to $\dot{\bu}$.
Both rows correspond to a negative event.}
\label{fig:DVSContrastAtPointSeveral}
\end{figure*}

Assuming constant illumination and independence of the observations, each event $e_{n}=\langle u_{n},v_{n},t_{n},p_{n}\rangle$ is caused by a brightness change at pixel~$\bp_{n}=(u_{n},v_{n})^{\top}$, depending on both the DVS state~$\x_{n}\equiv\x(t_{n})$ \emph{and} the map~$\cM$.
Thus, a more rigorous description than~\eqref{eq:MeasurementEqImplicit} is $\q=\q(\z_{n},\x_{n};\cM)$ because an event is an observation of some map point. 
Letting $\g$ be a shorthand notation for the spatial gradient $\nabla_{\bu}\log(\Int)$ in~\eqref{eq:IncrementLogI_PlanarPerspProj}, 
we define the implicit function $\q$ as the difference between the absolute contrast~\eqref{eq:ContrastDef} and the nominal threshold, $\q=|\Delta\log(I)|-C$. 
Substituting $|y|=y\,\sgn(y)$ for $y=\Delta\log(I)$ and replacing $\sgn(y)$ by the measured polarity~$p_{n}$, we use~\eqref{eq:IncrementLogI_PlanarPerspProj} to define
\begin{equation}
\q(\z_{n},\x_{n};\cM)=-p_{n}\left.\inner{\g}{\dot{\bu}}\right|_{(\bp_{n},\x_{n},\X_{n})}\Delta t_{n}-C,\label{eq:ImplicitMeasDef}
\end{equation}
where $\Delta t_{n}=t_{n}-t_{\text{prev}}$ is the time span since the previous event at the same location $\bp_{n}$, and the inner product between the gradient $\g$ and the motion field $\dot{\bu}$ depends on the event location $\bp_{n}$, its corresponding 3-D point $\X_{n}$ and the state $\x_{n}$. 
Specifically, $\g$ depends on the DVS pose only (but not on its velocity) via the perspective projection between the map $\X_{n}$ and point $\bp_{n}$,
whereas the motion field~\eqref{eq:ImageMotionFieldPlanar} depends on both the DVS pose (depth $Z$ of $\X_{n}$ with respect to the sensor) and velocities (twist coordinates).
The gradient $\g$ may be computed by taking the spatial derivatives of the predicted image intensities $I$ in a neighborhood of the current event location $\bp_{n}$, obtained through rendering the dense map $\cM$ according to the DVS pose in the current state.
Examples of the contrast function for positive and negative events are shown in Fig.~\ref{fig:DVSContrastAtPoint}.
Patches of $15\times 15$ pixels around the event location are displayed, but the local analysis of the generative event model is only reliable close to the center.
Fig.~\ref{fig:DVSContrastAtPointSeveral} reports the cases of moving edges parallel or almost perpendicular to the apparent motion, yielding largest and smallest absolute contrast, respectively.

\subsection{Recursive solution: Implicit EKF equations}

Once the system state and measurements equations have been designed, the update equations of the parameters of the posterior in the EKF are also determined. The recursive estimation carried out in the EKF is described by the equations in Algorithm~\ref{alg:IEKF}. 
We follow the notation in~\cite{Thrun05book} for the posteriors and their moments. 
The DVS pose tracking filter also assumes that an accurate estimate of the initial configuration, with relatively small uncertainties, is given $(\bmu_{0},\Sigma_{0})$. 
Let us further explain the steps of Algorithm~\ref{alg:IEKF}.

\begin{algorithm*}[t]
\begin{centering}
\begin{tabular}{ll}
1. Mean state (pred.) & $\bar{\bmu}_{n}=\f(\bmu_{n-1},\bw_{n})$\tabularnewline
2. Error covar. (pred.) & $\bar{\Sigma}_{n}=F_{n}\Sigma_{n-1}F_{n}^{\top}+L_{n}Q_{n}^{\bw}L_{n}^{\top}$, with $F_{n},L_{n}$ Jacobians of $\f$.\tabularnewline
3. Innovation & $\bnu_{n}=-\q(\z_{n},\bar{\bmu}_{n})$\tabularnewline
4. Innovation covar. & $S_{n}=H_{n}\bar{\Sigma}_{n}H_{n}^{\top}+R_{n}$, with $H_{n}$ and $R_n$ given by the Jacobians of $\q$.\tabularnewline
5. Kalman gain & $K_{n}=\bar{\Sigma}_{n}H_{n}^{\top}S_{n}^{-1}$ \tabularnewline
6. Mean state & $\bmu_{n}=\bar{\bmu}_{n}+K_{n}\bnu_{n}$\tabularnewline
7. Error covar. & $\Sigma_{n}=(I-K_{n}H{}_{n})\bar{\Sigma}_{n}(I-K_{n}H{}_{n})^{\top}+K_{n}R_{n}K_{n}^{\top}$\tabularnewline
\end{tabular}
\par\end{centering}
\protect\caption{\label{alg:IEKF}Extended Kalman Filter (EKF) equations for one iteration, $(\protect\bmu_{n-1},\Sigma_{n-1})\to(\protect\bmu_{n},\Sigma_{n})$, with \emph{implicit} measurement function $\protect\q$.}
\end{algorithm*}

\paragraph{Prediction.}

In this step, the projection of the posterior $\bel_{n-1}\sim\cN(\bmu_{n-1},\Sigma_{n-1})$ through the kinematic model~\eqref{eq:StateEqPartitioned} gives the predicted posterior $\overline{\bel}_{n}\sim\cN(\bar{\bmu}_{n},\bar{\Sigma}_{n})$ before incorporating the measurement. 
The state mean and error covariance are predicted according to lines (1)-(2) in Algorithm~\ref{alg:IEKF}. 
Uncertainty is propagated through the system by means of the Jacobians of~\eqref{eq:StateEqPartitioned}, 
$F_{n}=\partial\f/\partial\x_{n-1}$, $L_{n}=\partial\f/\partial\bw_{n}$, 
evaluated at the current best estimate, $(\bmu_{n-1},\bw_{n})$.

\paragraph{Correction.}

This is the data assimilation step, where the predicted posterior $\overline{\bel}_{n}\sim\cN(\bar{\bmu}_{n},\bar{\Sigma}_{n})$ is combined with the measurement $\z_{n}$ to yield the updated posterior $\bel_{n}\sim\cN(\bmu_{n},\Sigma_{n})$. 
The state mean and error covariance are corrected according to lines (3)-(7) in Algorithm~\ref{alg:IEKF}.
Events from the DVS are fed to the generative sensor equation~\eqref{eq:ImplicitMeasDef} to produce a residual that drives the update of the filter variables.
With regard to Figs.~\ref{fig:DVSContrastAtPoint}d and~\ref{fig:DVSContrastAtPointSeveral}d, the correction step changes the state such that the likelihood at the event position increases (white region).
The innovation process and its covariance (lines (3)-(4) in Algorithm~\ref{alg:IEKF}) are obtained by linearization of the implicit measurement function~\eqref{eq:ImplicitMeasDef} around the current best estimate, $(\z_{n},\bar{\bmu}_{n})$ (see~\cite{Zhang92IEKF,Soatto93IEKF}).
Uncertainty is corrected in the system (up to first order) by means of the Jacobians of~\eqref{eq:ImplicitMeasDef} (evaluated at $(\z_{n},\bar{\bmu}_{n};\cM)$),
$H_{n}=\partial\q/\partial\x_n$, $D_{n}=\partial\q/\partial\z_n$, with covariance of the measurement noise~\cite{Zhang92IEKF}
$R_{n}\coloneqq D_{n}\Cov_{n}^{\eeta}D_{n}^{\top}$.
Since $\q$ is a real value, both the noise and the innovation covariances ($R_{n}$ and $S_{n}$) are scalars.

\subsubsection{Data association}

An additional advantage of our approach is that there is no data association like in the classical localization problem (associating predicted measurements to actual ones), thus removing a challenging sub-problem and common source of brittleness in localization and mapping with the EKF.
This is a consequence of using a dense map (as opposed to a set of isolated landmarks) to represent the scene and to design a measurement equation~\eqref{eq:ImplicitMeasDef} that exploits such a representation. 
There is no data association problem because a correspondence between the event location and a map point $\bp_{n}\leftrightarrow\X_{n}$ will always exist, and it can be computed via ray-tracing.
The errors caused by a mismatch between the true surface point $\bar{\X}_{n}$ that triggered the event and the predicted one $\X_{n}$ are implicitly taken into account in the EKF via the innovation~\eqref{eq:ImplicitMeasDef} and its covariance. 
For example, the value of the gradient $\g$ in the neighborhood of the event will change (with some degree of smoothness) and if the predicted value does not yield the triggering of an event, the EKF adjusts the state parameters so that a different surface point $\X_{n}$ will be more likely to trigger the observed event. 
There is no need to artificially search for a 3-D point, close to the predicted one, that better explains the event.

\section{Experiments}
\label{sec:experiments}

\subsection{Synthetic data}
\begin{figure*}
\begin{center}
\begin{tabular}{ccc}
\includegraphics[width=0.305\linewidth]{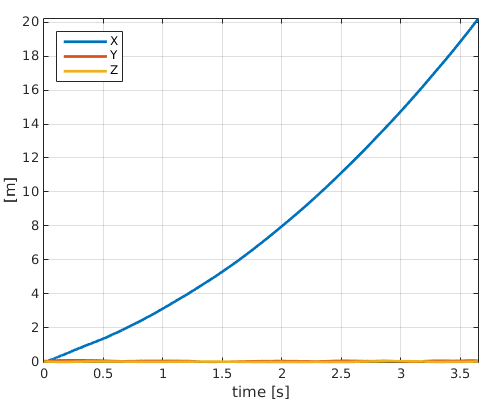} &
\includegraphics[width=0.305\linewidth]{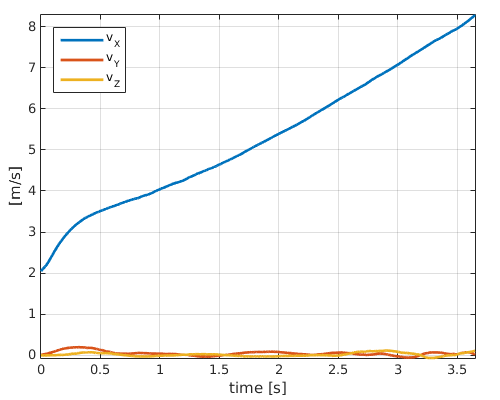} &
\includegraphics[width=0.33\linewidth]{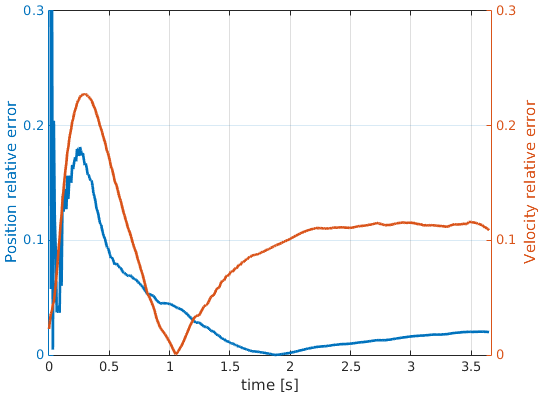}\\
(a)&(b)&(c)
\end{tabular}
\end{center}
\vspace{-2ex}
\caption{Constant acceleration experiment.
(a) Estimated position.
(b) Estimated velocity.
(c) Relative errors in position and velocity between simulated trajectory and estimated one.
}
\label{fig:ExpConstAccel}
\end{figure*}

The proposed method was tested with synthetic and real data.
The synthetic data was generated using computer graphics software (Blender\footnote{https://www.blender.org/}) to render images of a given map along a specified trajectory.
Adjacent images were subtracted, thresholded and randomly sampled to simulate the events generated by a DVS.
We chose a pinhole camera model with intrinsics identical to the ones of a lens from the real experiments: 2.6 mm lens for a 1/3'' sensor.
A linear trajectory with constant acceleration was simulated.
Results are reported in Fig.~\ref{fig:ExpConstAccel}.
Groups of 500 events every 8 ms were generated between adjacent images.
The algorithm processed 230k events.
This experiment validated the measurement function~\eqref{eq:ImplicitMeasDef} since the kinematic model~\eqref{eq:StateEqPartitioned} alone cannot predict the DVS motion.
The results show that the filter successfully estimated the DVS pose and velocity, with small relative errors (Fig.~\ref{fig:ExpConstAccel}c).

\subsection{Real data}
\begin{figure*}
\begin{center}
\begin{tabular}{ccc}
\;\includegraphics[width=0.252\linewidth]{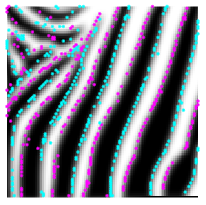} & 
\hspace{7mm}\includegraphics[width=0.33\linewidth]{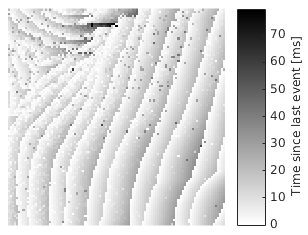} &
\includegraphics[width=0.23\linewidth]{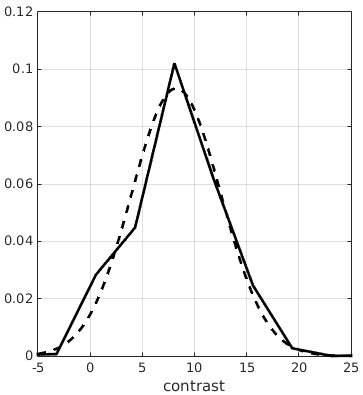}\\
(a)&(b)&(c)\\
\hspace{-2mm}\includegraphics[width=0.27\linewidth]{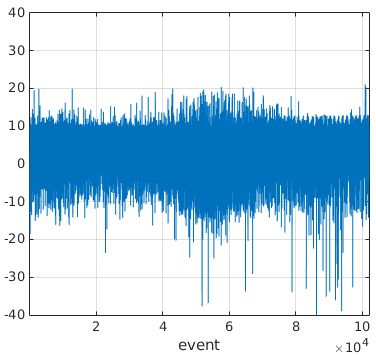} &
\hspace{-5mm}\includegraphics[width=0.31\linewidth]{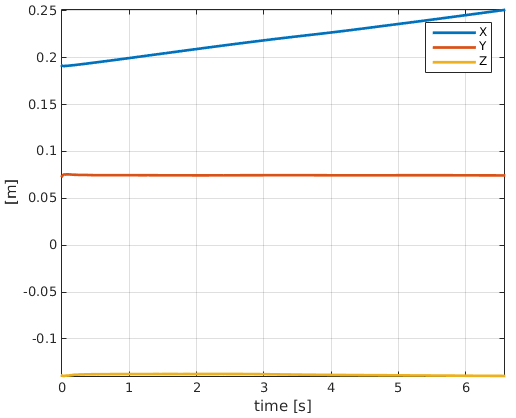} &
\includegraphics[width=0.315\linewidth]{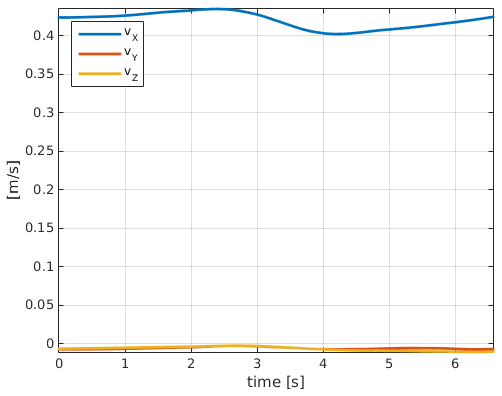}\\
(d)&(e)&(f)
\end{tabular}
\end{center}
\vspace{-1ex}
\caption{Experiment with approximately constant velocity motion.
(a) Visualization of a few events from the DVS (positive events in cyan, negative events in magenta) used for filter initialization, 
overlaid on the rendered map.
(b) Time since the last event at each pixel ($\Delta t_n$ in~\eqref{eq:ImplicitMeasDef})
(c) Normalized histogram of the absolute contrast in~\eqref{eq:ImplicitMeasDef} (solid line) and Gaussian fit (dashed line) (cf. Fig.~\ref{fig:DVSContrast2008Composite}c).
The mode of the Gaussian corresponds to the threshold $C$.
(d) Innovations sequence $\bnu_n$.
Estimated position (e) and velocity (f) of the event-based camera.}
\label{fig:ExpLinMotion}
\end{figure*}

For the experiment with real data, 
we mounted the DVS on a model train that runs on a straight track with constant velocity. 
The DVS faced sideways and observed a planar scene at a constant distance. 
The scene contains a pattern of complex black and white stripes and a set of circles at known locations; the latter were used for extrinsic calibration. 
The DVS was intrinsically calibrated using standard camera calibration techniques on the imaged points detected from the projection of an array of blinking LEDS placed in a checkerboard configuration.
Horizontal edges are parallel to the apparent motion, and, consequently do not trigger events.
The intensities of the map were smoothed to provide non-zero gradients in the regions near sharp edges that generate events, hence to smooth the response of the contrast function~\eqref{eq:ImplicitMeasDef} and the corresponding likelihood in such regions.
Fig.~\ref{fig:ExpLinMotion} reports some of the results of this experiment.
Fig.~\ref{fig:ExpLinMotion}b shows, for a few hundreds of events (Fig.~\ref{fig:ExpLinMotion}a), the measured absolute contrast $|\Delta \tilde{I}| \approx -p_n \inner{\g}{\dot{\bu}}\Delta t_n$ used in the implicit measurement function~\eqref{eq:ImplicitMeasDef}.
Having the map intensities given in arbitrary units (log of gray levels) and lacking physical measurements of the incoming light that the DVS used to trigger the events, the threshold values in Fig.~\ref{fig:DVSContrast2008Composite}c ($\approx 0.2$) are not applicable to the map, and so a few events are used to estimate the threshold $C$ corresponding to the given map.
The filter processed about 100k events and successfully 
estimated the DVS pose and velocities of the DVS throughout the event stream. 
Figs.~\ref{fig:DVSContrastAtPoint} and~\ref{fig:DVSContrastAtPointSeveral} were also obtained from this experiment.

\section{Conclusion}
\label{sec:conclusion}

We have successfully developed an implicit EKF for event-based camera (DVS) localization based on the contrast residual~\eqref{eq:ImplicitMeasDef}, which provides a natural measure to define the likelihood of an event.
For this, we derived a generative event model that incorporates the physical characteristics of the DVS.
Our algorithm readily matches the asynchronous nature of the events and allows filter updates on an event-by-event basis.
An additional advantage of our approach is that the contrast residual naturally takes into account a dense map representation of the environment, removing the data-association sub-problem.
In future work, we plan to extend the developed method to event-based SLAM without additional sensing.

\balance


\end{document}